\documentclass[letterpaper]{article} %
\usepackage{aaai23}  %
\usepackage{times}  %
\usepackage{helvet}  %
\usepackage{courier}  %
\usepackage[hyphens]{url}  %
\usepackage{hyperref}
\usepackage{graphicx} %
\usepackage{natbib}  %
\usepackage{caption} %
\usepackage{algorithm}
\usepackage{algorithmic}

\usepackage{tabularx}
\usepackage{booktabs}
\usepackage{cuted}

\usepackage{newfloat}
\usepackage{listings}
\DeclareCaptionStyle{ruled}{labelfont=normalfont,labelsep=colon,strut=off} %
\floatstyle{ruled}
\newfloat{listing}{tb}{lst}{}
\floatname{listing}{Listing}
\title{Envisioning the Next-Gen Document Reader}
\author {
    Catherine Yeh\textsuperscript{\rm 1},
    Nedim Lipka\textsuperscript{\rm 2},
    Franck Dernoncourt\textsuperscript{\rm 2}
}
\affiliations {
    \textsuperscript{\rm 1} Harvard University\\
    \textsuperscript{\rm 2} Adobe Research\\
    catherineyeh@g.harvard.edu, lipka@adobe.com, franck.dernoncourt@adobe.com
}

\usepackage{bibentry}

\begin{document}

\maketitle

\begin{strip}
    \includegraphics[width=\textwidth]{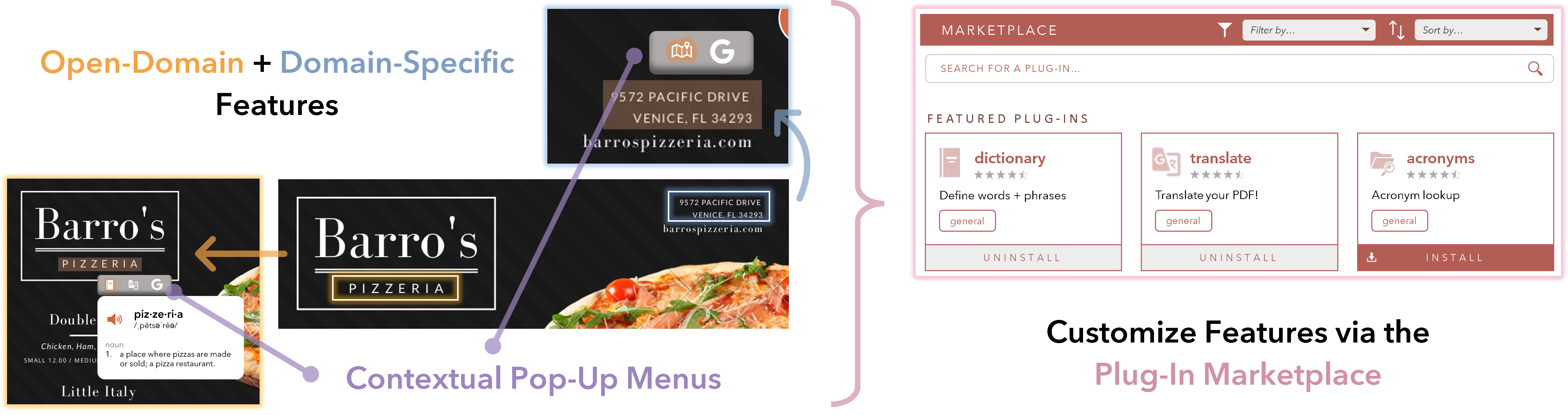}
    \captionof{figure}{Overview of our vision for the next-gen document reader. We propose transforming static, isolated documents into a trustworthy, connected, and interactive sources of information through open-domain and domain-specific features, contextual plug-in pop-up menus, and a centralized plug-in marketplace. (Menu source: \href{https://www.pdf-archive.com/2020/05/15/white-black-pizza-italian-stripes-restaurant-menu/}{PDF Archive website})}
    \label{fig:overview}
\end{strip}

\begin{abstract}
People read digital documents on a daily basis to share, exchange, and understand information in electronic settings. However, current document readers create a static, isolated reading experience, which does not support users' goals of gaining more knowledge and performing additional tasks through document interaction. In this work, we present our vision for the next-gen document reader that strives to enhance user understanding and create a more connected, trustworthy information experience. We describe 18 NLP-powered features to add to existing document readers and propose a novel plug-in marketplace that allows users to further customize their reading experience, as demonstrated through 3 exploratory UI prototypes available at: \href{https://github.com/catherinesyeh/nextgen-prototypes}{github.com/catherinesyeh/nextgen-prototypes}.

\end{abstract}

\section{Introduction}
Digital documents (e.g., portable document format (PDF) files or Word documents) are a popular format for sharing, exchanging, and understanding information in electronic settings. 
Reading such documents is an integral part of countless people's daily routines, 
and many choose to engage with these files through document readers such as \href{https://www.adobe.com/acrobat/pdf-reader.html}{\textit{Adobe Acrobat}}, \href{https://www.foxit.com/pdf-reader/}{\textit{Foxit}}, and \href{https://www.sumatrapdfreader.org/free-pdf-reader}{\textit{Sumatra PDF}}. However, with these current applications, document reading can feel relatively static and isolated, as the reading experience is usually confined to within the document reader itself. Additionally, there is typically not much interaction between the user and the information they are reading when scrolling through a digital document.

This presents a problem as documents themselves are usually not the end goal for users. Rather, they represent a starting point for people to gain more knowledge or perform additional actions. Thus, in this work, we present our \textbf{vision for the next-gen document reader} that strives to better support users in achieving their goals through harnessing the power of natural language processing (NLP). We design this next-gen document reader to 1) enhance user understanding of digital files and 2) transform currently static, isolated documents into connected, trustworthy, and interactive sources of information. 

The key contributions of our work include:

\begin{itemize}
    \item A set of proposed NLP-powered plug-ins to add to existing document readers toward enhancing human-document interaction, including 12 \textbf{open-domain} and 6 \textbf{domain-specific} features.
    \item A preliminary vision for a centralized \textbf{plug-in marketplace} that would allow further customization of the user experience in document readers and feature development to be outsourced.
    \item 3 exploratory \textbf{UI prototypes} illustrating a subset of features and the plug-in marketplace proposed for the next-gen document reader (\href{https://github.com/catherinesyeh/nextgen-prototypes}{github.com/catherinesyeh/nextgen-prototypes}).
\end{itemize}

\section{Related Work}
While older document readers such as \textit{Adobe Acrobat}, \textit{Foxit}, and \textit{Sumatra PDF} tend to only support static, in-document features, recent NLP efforts are beginning to explore the possibility of creating a more connected, trustworthy information experience for users.

For example, \textit{ScholarPhi} \citep{head2021augmenting} strives to improve the readability of scientific papers by creating an augmented reading interface with features such as position-sensitive definitions, a decluttering filter, and an automatically generated glossary for the important terms and symbols. Similarly, \textit{Paper Plain} \citep{august2022paper} is an interactive interface that aims to make medical research papers more accessible with its definition feature, section gists, and Q \& A passages. \textit{Scim} \citep{fok2022scim} is another AI-augmented document reader that helps researchers skim scientific papers by automatically identifying, classifying, and highlighting salient sentences. 

\href{https://sioyek.info/}{\textit{Sioyek}}~\cite{Sioyek}, a document viewer designed for reading technical books and research papers, has some interesting features such as smart jump for references and figures, searchable bookmarks, and portals to display linked information in a separate window. \href{https://www.explainpaper.com/}{\textit{Explainpaper}} is a novel AI-powered reading interface for reading academic papers as well, offering live explanations to users upon highlighting sections of text and an interactive Q \& A feature. However, these works are currently very limited in their features and scope.

\begin{figure}[ht]
    \centering
    \includegraphics[width=\columnwidth]{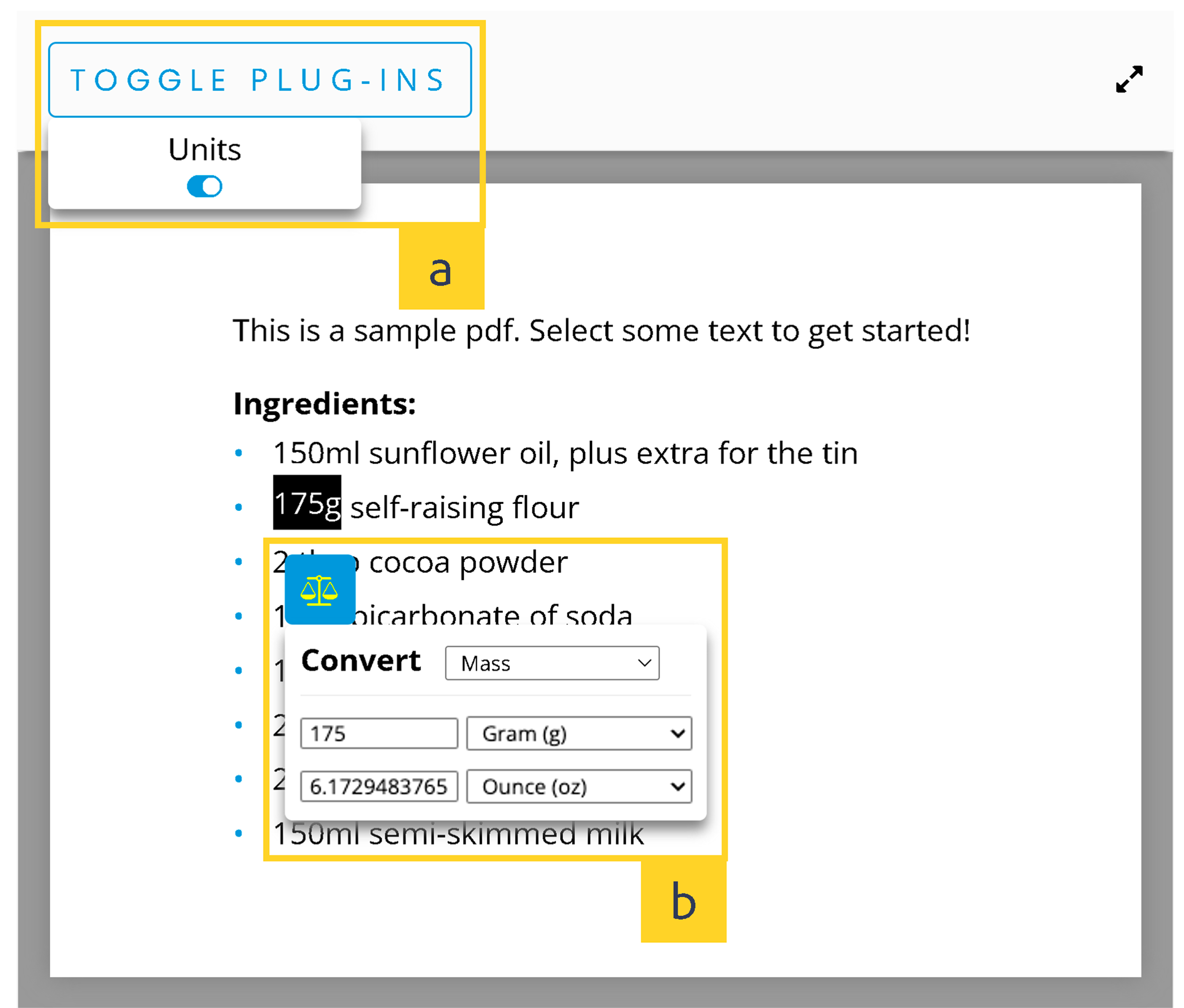}
    \caption{UI prototype demonstrating our contextual pop-up menu concept. Users can a) toggle installed plug-ins on or off and when active, b) the corresponding plug-in tooltip only appears if relevant text is selected. \textit{(Recipe source: BBC Good Food)}}
    \label{fig:menu}
\end{figure}

\section{Vision}
Our vision for the next-gen document reader includes the following components as illustrated in Figure \ref{fig:overview}:
\begin{itemize}
    \item A set of \textbf{open-domain} features that can enhance the document reading experience for various document types,
    \item A set of features that are more \textbf{domain-specific}, and
    \item A centralized \textbf{plug-in marketplace} that would allow users to further customize document readers with additional features.
\end{itemize}

Throughout this paper, we use plug-in and feature synonymously to mean any software add-on that serves to extend the core functionality of a static document reader. Once a plug-in is installed, it could be toggled on/off by users and when active, the plug-in could be accessed through a \textbf{contextual pop-up menu}. Figure \ref{fig:menu} illustrates how this type of menu could work with a sample PDF cake recipe. In this UI prototype, the unit conversion plug-in is toggled on, but the corresponding tooltip icon only appears if relevant text is selected (i.e., text containing numerical values). Ideally, the document reader would also automatically identify the correct unit of measurement selected by the user and auto-populate this information into the pop-up conversion tool.

In the following sections, we provide more details about our proposed features and plug-in marketplace for the next-gen document reader.

\begin{table}[h]
\caption{List of open-domain and domain-specific features proposed for the next-gen document reader}
\label{tab:features}
{\renewcommand{\arraystretch}{1.2}
\begin{tabularx}{\columnwidth}{p{0.5\columnwidth}p{0.5\columnwidth}}
\hline
\textbf{Open-Domain}              & \textbf{}          \\ \hline
Definitions                       & Equation Exporting      \\
Acronyms/Abbreviations            & Speed Reading \\
Unit Conversions                  & Sentiment Analysis     \\
Translations                      & Form Auto-Fill      \\
Spelling Changes                  & Scholar Notes  \\
Table Copying                     & Shared Commenting  \vspace{2mm}   \\ \hline
\textbf{Domain-Specific}          &                    \\ \hline
Linkifying Known Entities          & Smart Jumps  \\
Linkifying Relevant Content      & Citation Warnings        \\
Action Tasks                      & Portals            \\ \hline
\end{tabularx}
}
\end{table}

\subsection{Features}
To begin the design process, we brainstormed features that would be helpful to add to static document readers such as \textit{Acrobat} or \textit{Foxit}, focusing on features that can leverage NLP. During this stage, we surveyed the literature and investigated existing plug-ins supported by newer document viewers \citep{august2022paper, fok2022scim, head2021augmenting} as described in the Related Work section. Some ideas were also contributed by peers and collaborators. 

This process resulted in 26 potential feature suggestions, which we narrowed down to 18 based on feasibility of implementation. These ideas were then categorized by domain type, ultimately yielding 12 \textbf{open-domain} and 6 \textbf{domain-specific} features.
A list of all proposed features is included in Table \ref{tab:features}. Selected features are highlighted below using our second UI prototype of a PDF pizza restaurant menu (as previewed in Figure \ref{fig:overview}).

\subsubsection{Open-Domain Features}
\begin{figure}[h]
    \centering
    \includegraphics[width=\columnwidth]{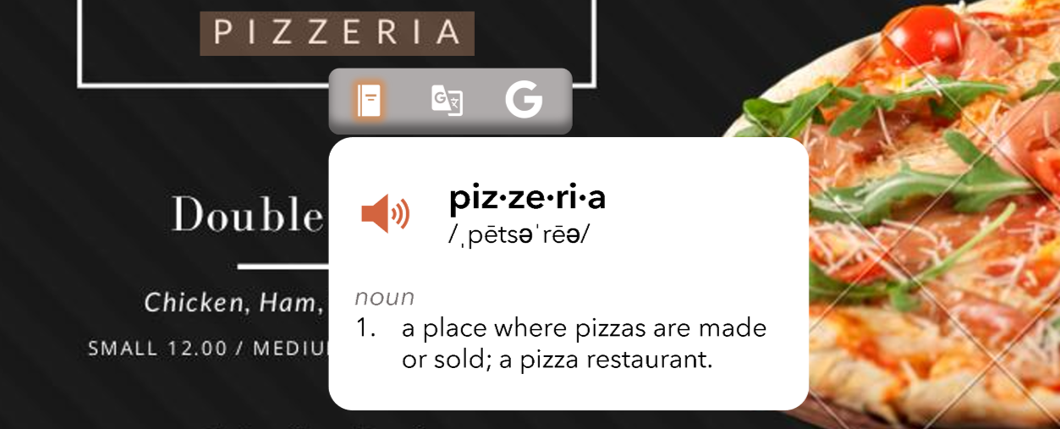}
    \caption{Example definition feature}
    \label{fig:definition}
\end{figure}

When a single word is highlighted, document readers could show users potential \textbf{definitions} of the term (Figure \ref{fig:definition}), similar to \cite{august2022paper, head2021augmenting}. The displayed definitions could be retrieved from the document itself~\cite{veyseh2020joint,spala-etal-2020-semeval}, from online sources as in \cite{august2022paper}, or crowd-sourced. %

\begin{figure}[h]
    \centering
    \includegraphics[width=\columnwidth]{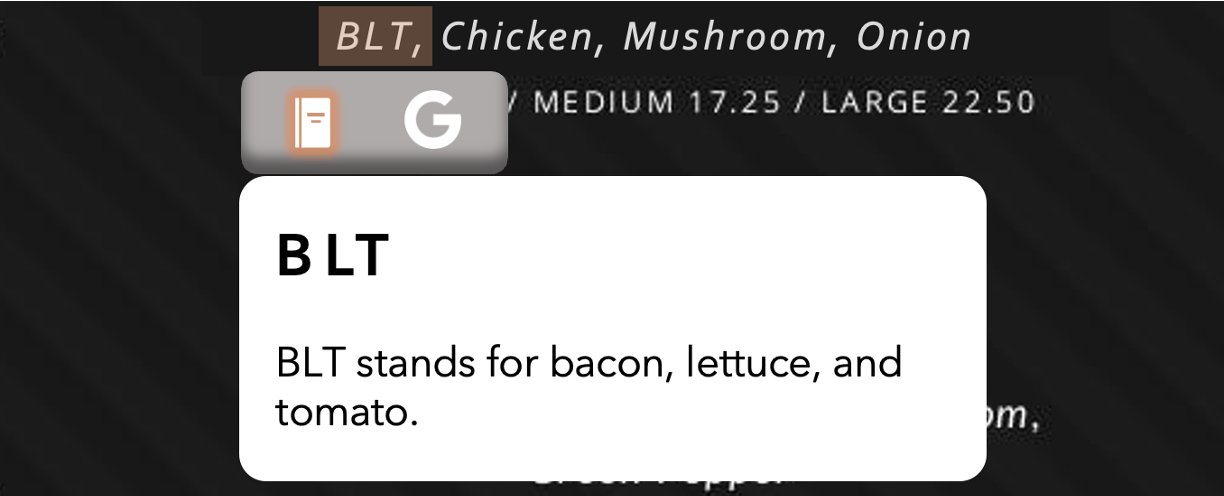}
    \caption{Example acronym feature}
    \label{fig:acronym}
\end{figure}

Similarly, the long forms of \textbf{acronyms/abbreviations} could be shown to the user via pop-up tool tips (Figure \ref{fig:acronym}). If the acronym is defined elsewhere in the paper, we could take a similar approach to \cite{veyseh2020acronym,pouran-ben-veyseh-etal-2020-acronym,veyseh2022acronym} for extracting definitions; otherwise, retrieving it from online sources is also possible. A list of key definitions and acronyms could be included at the beginning or the end of the document as well, following \cite{pouran-ben-veyseh-etal-2021-maddog}. The definition of math symbols could also be extracted from the text~\cite{lai-etal-2022-semeval,lai2022symlink}.

\begin{figure}[ht]
    \centering
    \includegraphics[width=\columnwidth]{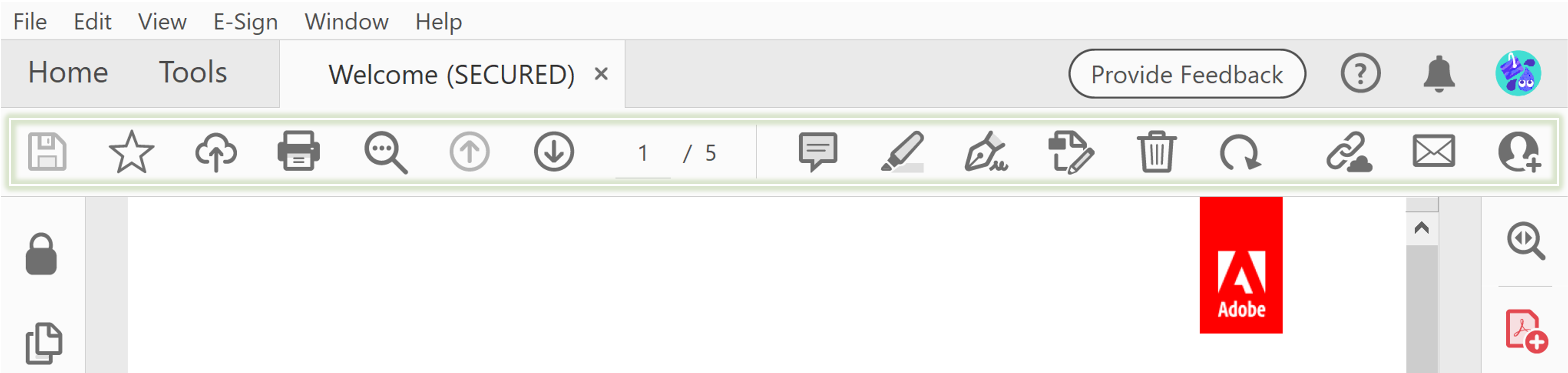}
    \caption{Main toolbar in Adobe Acrobat}
    \label{fig:toolbar}
\end{figure}

\textbf{Unit conversions} could be either a case-by-case basis (i.e., users highlight specific numbers to convert like definitions/acronyms) or document-level (i.e., document reader automatically converts all units at once). Figure \ref{fig:menu} shows the former option, assuming that unit selection is embedded inside of the plug-in pop-up. Allowing users to select their unit of choice via the main document reader toolbar (see Figure \ref{fig:toolbar}) would be another possibility.

\begin{figure}[h]
    \centering
    \includegraphics[width=\columnwidth]{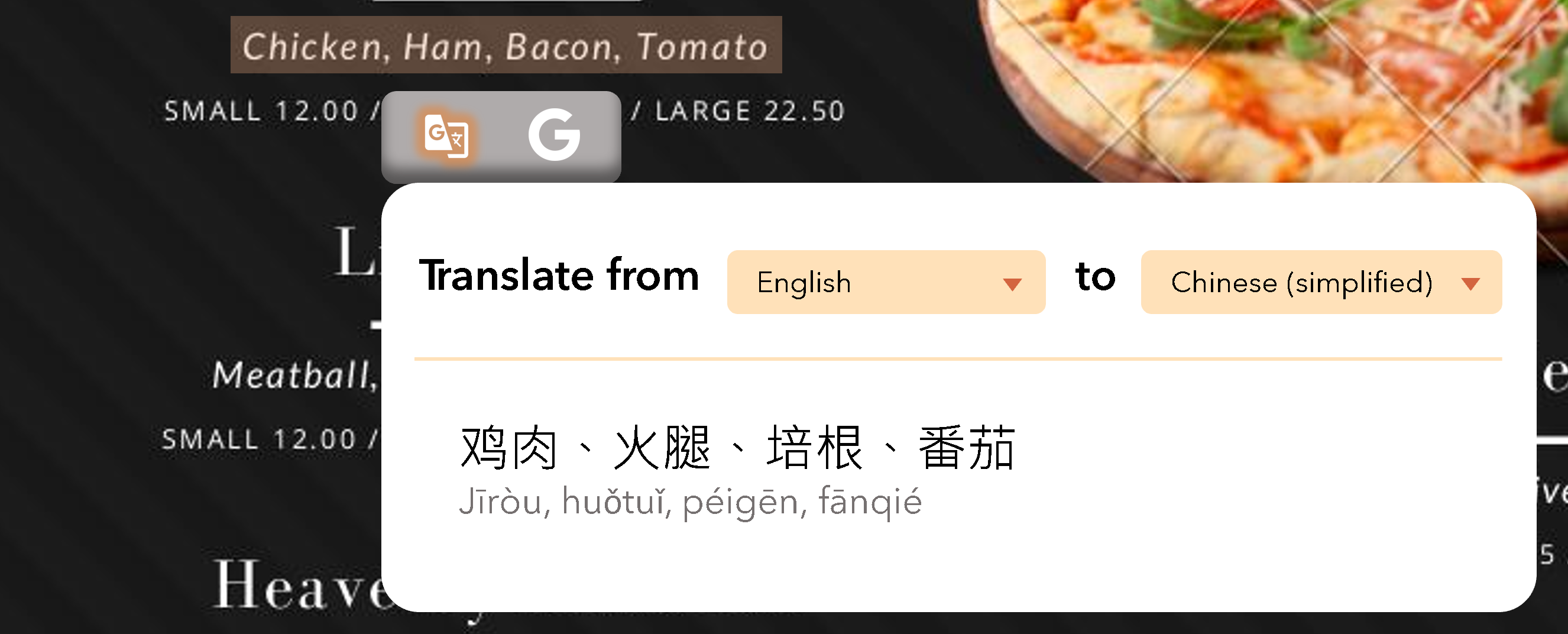}
    \caption{Example translation feature}
    \label{fig:translate}
\end{figure}

As with unit conversions, there could be a toolbar option at the top of document readers that would allow users to choose the language they want to read the document in. Alternatively, 
\textbf{translations} could be performed on a more case-by-case basis. The former could be similar to the \href{https://chrome.google.com/webstore/detail/google-translate/aapbdbdomjkkjkaonfhkkikfgjllcleb?hl=en}{Google Translate browser extension}. Figure \ref{fig:translate} illustrates the latter option. Related to translation is the idea of automatically suggesting \textbf{spelling changes} based on the current document language. For example, the spellings in a document could automatically be converted from American to British English (e.g., color $\rightarrow$ colour).

Document readers could also provide users with the capability of directly copying tables from files like PDFs into Microsoft Word, Excel, Markdown, etc. to further manipulate, share or analyze. This \textbf{table copying} feature would reduce the need for hand transcribing and combining data from multiple tables in digital files. There could also be a selection hierarchy, allowing users to select specific parts of a table (e.g., a single cell, row, column, etc.) or the whole table itself. 
A version of this feature is included in the \href{https://developer.adobe.com/document-services/apis/pdf-extract/}{\textit{Adobe PDF Extract API}}, but currently, tables can only be exported to CSV formats, so there is room for extension and it is still missing from all the PDF readers we surveyed. Similarly, an \textbf{equation exporting} plug-in could allow users to export math equations present in digital documents to their corresponding \href{https://www.latex-project.org/}{LaTeX} formulas so they are directly editable. Implementing this feature would be possible using image-to-latex algorithms~\cite{deng2017image,10.1007/978-3-030-10925-7_2,zhang2020tree}.

Another way to enhance the document reading experience could be including a \textbf{speed-reading} plug-in that would allow allow users to customize the speed at which they read text in document readers, similar to the service offered by \href{https://spritz.com/}{\textit{Spritz}}. Additionally, a \textbf{sentiment analysis} feature could allow users to assess sentiment at the document level and potentially at the sentence level as well. Sentiment classification would be useful for a wide variety of document types, particularly when it is beneficial to understand a document's tone/attitude. Some approaches for document-level sentiment analysis have been proposed by \cite{ito2020contextual,rhanoui2019cnn}.

For certain documents such as history books, scientific papers, and poems, reading applications could also offer \textbf{scholar notes}. As an example, the document reader could include a critic's analysis of a text (e.g., in the sidebar) and/or their annotations throughout the file as comments that the user could view. The notes could be distributed via a marketplace, and some of them could be set as paid access only if monetization is of interest. Users may be willing to pay to get access to the meta-information given by a scholar in the field to better understand the text itself, its historical context, the equations, potential errors, the author's mindset at the time of the writing, and so on. A related feature idea is allowing users to leave \textbf{shared comments} in digital documents. For example, users could highlight a sentence or figure and then create a thread for further discussion (e.g., asking a question, offering clarification, etc.), which could open up in a sidebar. 

These human-in-the-loop features could help make up for the imperfections and the limitations of other AI-powered document plug-ins. However, the main challenges with implementing such features would be moderating/filtering the user content and respecting users' privacy (e.g., we do not want a user to mistakenly post their comments as public if they did not intend too).

\subsubsection{Domain-Specific Features}
\begin{figure}[h]
    \centering
    \includegraphics[width=\columnwidth]{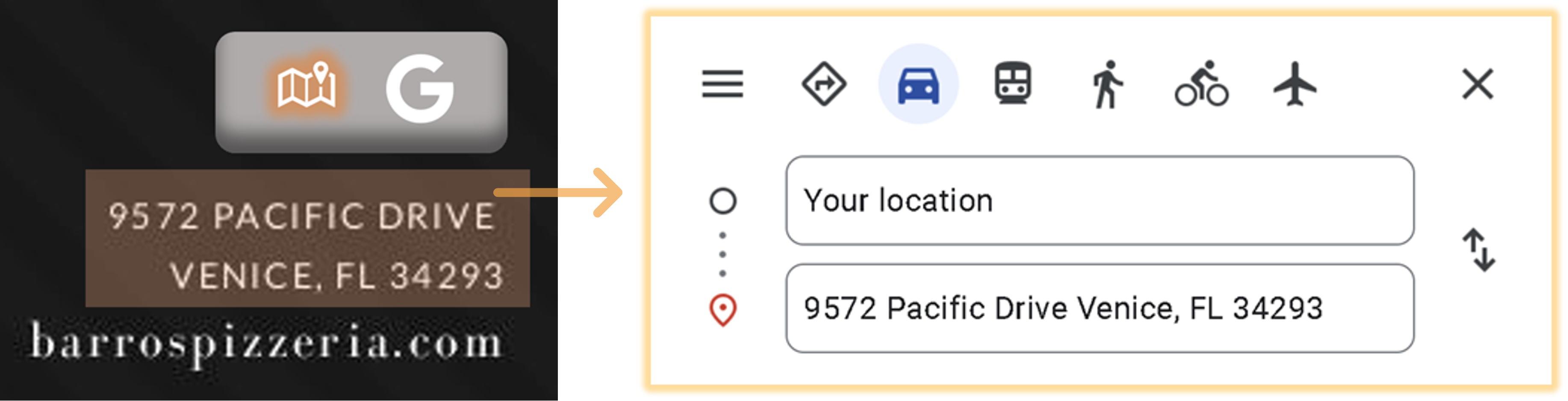}
    \caption{Example address linkifying feature}
    \label{fig:address}
\end{figure}

Document readers could also incorporate domain-specific features such as \textbf{linking text to known entities}. For example, addresses or business names could be automatically linked to Google Maps and phone numbers could be linked to an app/website for further action, as Google Chrome or Android currently does. The former is illustrated in Figure \ref{fig:address}.
Other ideas include linking protein names to the \href{https://www.rcsb.org/}{Protein Data Bank} for biology documents, linking references to their Google Scholar entry in scholarly articles, or linking ticker symbols to their Yahoo Finance pages (e.g., \href{https://finance.yahoo.com/quote/ADBE}{finance.yahoo.com/quote/ADBE} $\rightarrow$ ADBE) for finance documents.

\begin{figure}[ht]
    \centering
    \includegraphics[width=\columnwidth]{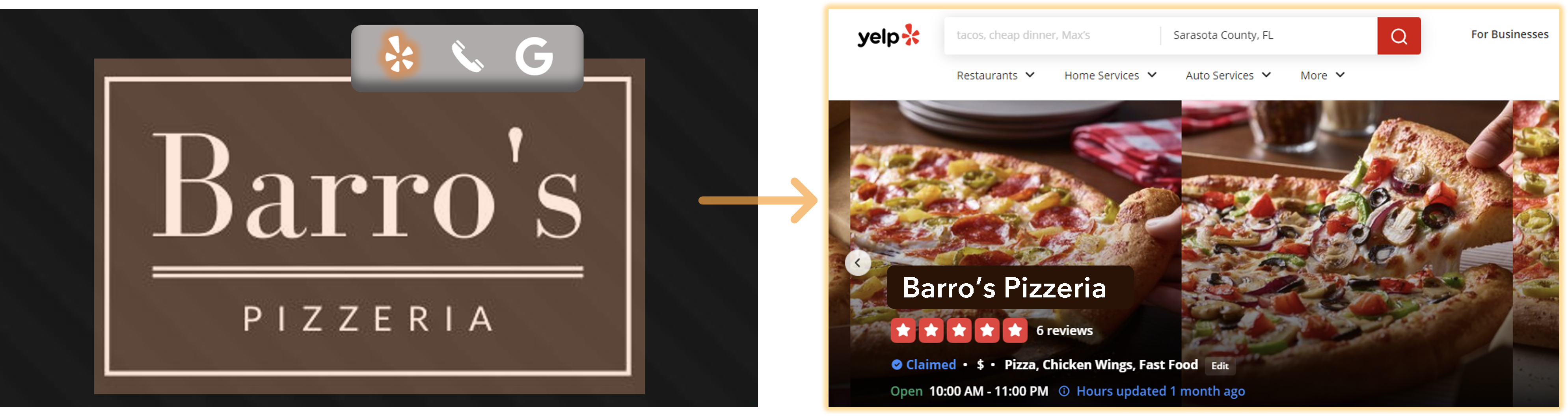}
    \caption{Example restaurant linkifying feature}
    \label{fig:yelp}
\end{figure}

Similarly, document readers could \textbf{link text to relevant content}. This is a trickier task than linkifying known entities, but the required extrapolation may be feasible in certain cases. For instance, if a file is identified as a restaurant menu, the document reader could link to the corresponding Yelp or Google reviews page so users could see more pictures/reviews of different items (Figure \ref{fig:yelp}). Or, if a movie title is identified inside a document, links to available movie times or streaming platforms could be generated. Another possibility would be searching selected keywords/phrases in a search engine or e-commerce website (e.g., Amazon, Alibaba, etc.) to see related products; \href{http://api-doc.axesso.de/##api-Amazon}{\textit{Axesso Amazon API}} has implemented one such keyword search feature. Ultimately, this feature could be similar to how YouTube recommends products based on the videos a user watches. 

Additionally, we could identify and create \textbf{action tasks} that users could complete within the document reader itself. For example, if a date/deadline is identified in the text, users could be given the option to add it to their calendar. Similarly, if a payment is mentioned, users could have the option to pay directly inside of the document. Or, if there is language such as ``You should notify X...'' or ``Please reach out to Y...'' in a digital file, it might be helpful to give users the ability to send messages/emails from the document reader as well. In general, these tasks could be accessed via pop-up icons throughout the document, but there could also be an overall list on the sidebar, for instance.

For documents like academic papers or textbooks, a \textbf{smart jump} feature (terminology from \textit{Sioyek}) could be offered that allows users to jump to any referenced figure or reference in the document, even if links are not explicitly provided. A similar feature has already been implemented by \textit{Sioyek}. 
Currently, \textit{Sioyek}'s smart jump feature automatically links references to their Google Scholar page as well, connecting to our idea for linkifying known entities described above. Another related idea would be including a \textbf{citation warning} feature that displays a warning to users when a citation in a document has been retracted~\cite{teixeira2017some,bolland2022citation}.

\begin{figure}[ht]
    \centering
    \includegraphics[width=\columnwidth]{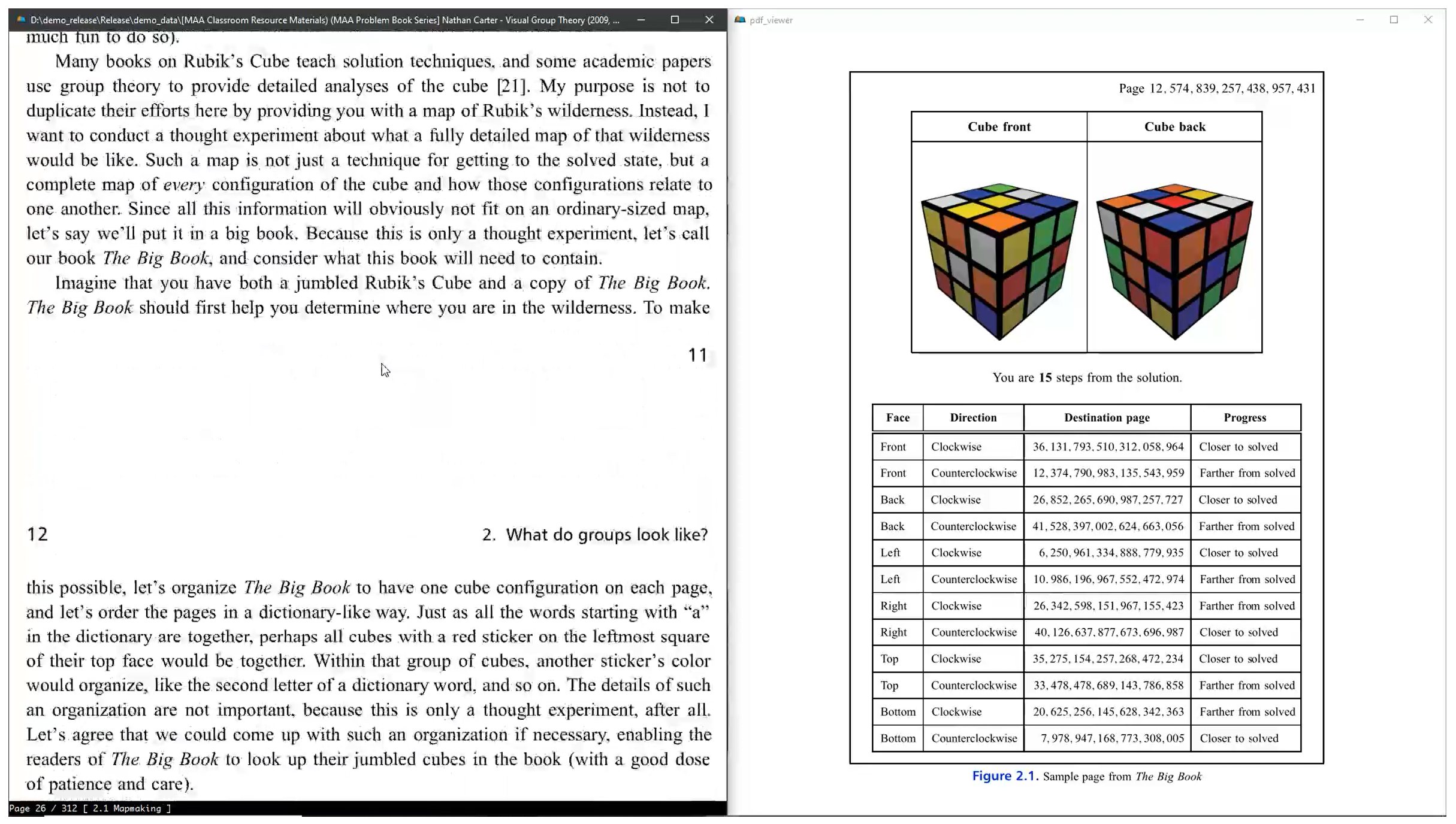}
    \caption{Example portal feature \textit{(Source: Sioyek)}}
    \label{fig:portals}
\end{figure}

\begin{figure*}[t]
    \includegraphics[width=\textwidth]{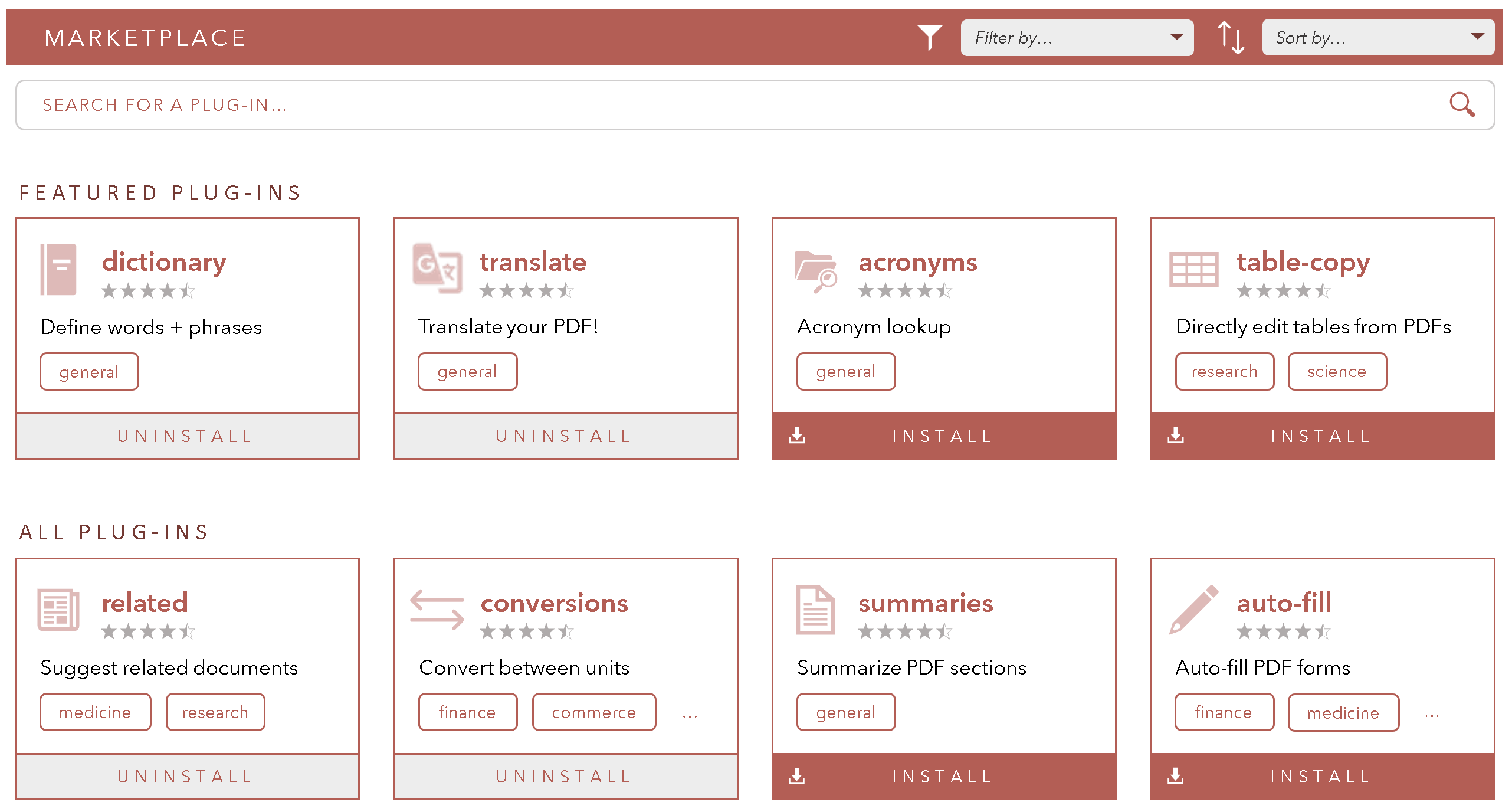}
    \captionof{figure}{UI Prototype of our proposed plug-in marketplace for the next-gen document reader. Users can discover new plug-ins via the search bar or featured section, filter or sort the results, and (un)install plug-ins at any time.}
    \label{fig:marketplace}
\end{figure*}

\begin{figure}[t]
    \centering
    \includegraphics[width=\columnwidth]{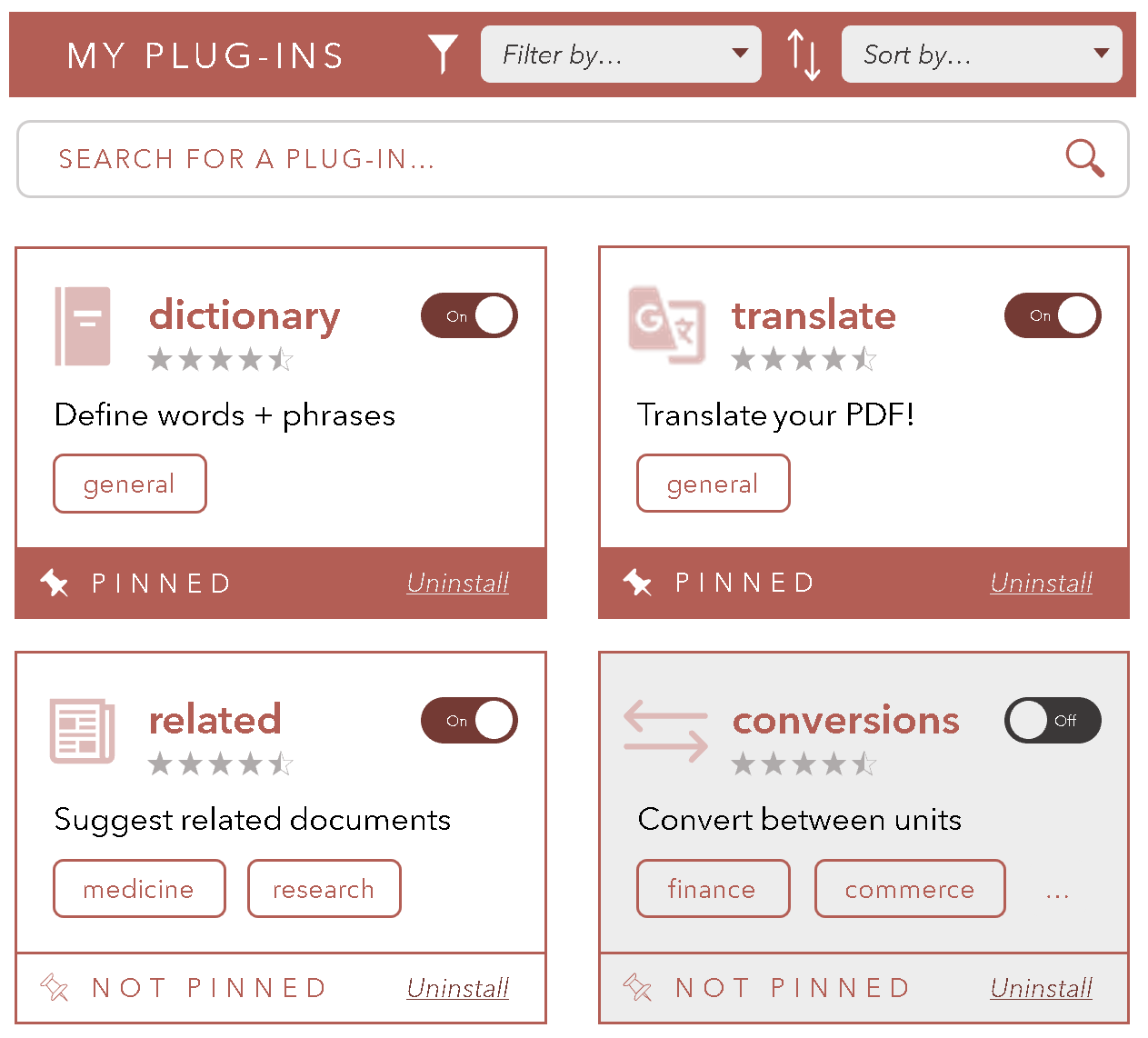}
    \caption{``My Plugins'' page inside plug-in marketplace}
    \label{fig:myplugins}
\end{figure}

One last domain-specific feature would be \textbf{portals} (terminology from \textit{Sioyek}). This feature would allow users to link figures to specific paragraph locations so they can view them simultaneously on a separate monitor/window, as implemented by \textit{Sioyek's} portal feature (Figure \ref{fig:portals}). Such ``portals'' would enhance the reading experience by removing the need to scroll back and forth in a document to find the relevant figures for each section of text, which could especially be helpful for academic/scientific papers.

\subsection{Plug-in Marketplace}
In the previous sections, we describe many potential plug-ins for the next-gen document reader. However, the average user will not need all these features when engaging with digital files. Thus, to allow users to further customize their document reading experience and choose which features they want to use, we propose the creation of a centralized \textbf{plug-in marketplace}. This way, document readers could come with a few default plug-ins (e.g., definitions/acronyms or other open-domain features that could be useful for most document types and users) and users could add more via the marketplace if they wanted to. 

Having a plug-in marketplace would also prevent document readers from growing excessively in terms of size and computational requirements. Instead, the user would individually decide which plug-ins to install and use, just like in virtually all modern code and text editors. %

\subsubsection{Key Features}
An example of what such a marketplace might look like is illustrated in Figure \ref{fig:marketplace}. To our knowledge, there currently exist no plug-in marketplaces for document readers such as \textit{Adobe Acrobat}, \textit{Foxit}, and \textit{Sumatra PDF}. Consequently, many features we include in our UI prototype are inspired by the marketplaces for integrated development environments (IDEs) like \href{https://marketplace.visualstudio.com/}{\textit{Visual Studio}}, \href{https://plugins.jetbrains.com/}{\textit{IntelliJ}}, and \href{https://marketplace.eclipse.org/}{\textit{Eclipse}} as well as marketplaces for text editors like \href{https://melpa.org/#/}{\textit{Emacs}}, \href{https://notepad-plus-plus.org/resources/#plugins}{\textit{Notepad++}}, and \href{https://packagecontrol.io/}{\textit{Sublime Text}}.

On the main marketplace page (Figure \ref{fig:marketplace}), we envision a space where users can \textbf{discover} new features with the search bar or featured plug-ins section, which could highlight the newest or most popular plug-ins. Plug-ins could also be tagged (e.g., by domain) and reviewed to allow users to \textbf{filter} and \textbf{sort} the results. Inside the marketplace, users would have the option to \textbf{(un)install} plugins at any time.

Each user could also have their own ``My Plugins'' page inside the plug-in marketplace, as shown in Figure \ref{fig:myplugins}. On this page, users would again have the ability to search for, filter, sort, and uninstall plugins. In addition, users could \textbf{pin} their favorite plug-ins for easy access. Here, users would be able to globally \textbf{toggle plugins on/off} as well. There could also be an option to turn plugins on/off at a document level, illustrated in Figure \ref{fig:menu}a.

Another feature that could be added within the plug-in marketplace is a \textbf{feedback/feature request} page, where users could submit general feedback about the marketplace or propose new ideas for features to add. Along these lines, several IDEs/text editors include a community \textbf{forum} (e.g., \href{https://www.eclipse.org/forums/}{eclipse.org/forums}) where users can openly discuss topics and ask questions about different plug-ins, so implementing a similar discussion platform for document readers could also be valuable.

\section{Future Work}
This work represents a preliminary, exploratory vision for the next-gen document reader. The next steps include working toward concretizing our ideas and assessing the viability of implementation. Specifically, we hope to conduct formal \textbf{user studies} to collect additional feedback on our vision, further hone the proposed designs, and better understand which features would be most useful to end-users. Through these user studies, we may also generate additional ideas for potential document reader plugins.

Further out in the future, %
we could also consider more \textbf{complex features}. %
For example, a filtering option would help readers focus on only the most relevant parts of the document, similar to the declutter feature from \cite{august2022paper}. Similarly, fact-checking sentences and displaying a warning symbol next to text containing incorrect facts would be extremely valuable. Other complex features for consideration include summarization~\cite{cohan-etal-2018-discourse,lee2022factual}, section title generation~\cite{rizvi-etal-2019-margin,gehrmann-etal-2019-improving}, key sentence highlighting~\cite{spala-etal-2018-comparison,spala-etal-2018-web}, and question-answering~\cite{tran-etal-2020-explain,xiao-etal-2021-open}. These ideas are more challenging to realize at the moment and may not be mature enough to be released to the general public, but the recent progresses in large language models are making some of these features more achievable~\cite{goyal2022news,zhang2023benchmarking}.

\section{Conclusion}
In this paper, we present our \textbf{vision for the next-gen document reader} that will transform static, isolated documents into connected, trustworthy, interactive sources of information. This vision includes 12 \textbf{open-domain} and 6 \textbf{domain-specific} features powered by NLP, which can be accessed by the user through contextual plug-in pop-up menus while reading digital files. To allow users to customize their reading experiences with document readers, we also propose a centralized \textbf{plug-in marketplace} inspired by modern IDEs and text editors.
Next steps include conducting formal user studies to further hone our UI prototypes (\href{https://github.com/catherinesyeh/nextgen-prototypes}{github.com/catherinesyeh/nextgen-prototypes}) and vision, while also considering additional complex features to improve user-document interaction such as filtering or question-answering. 
We hope this work inspires and excites others about the future of document readers.

\bibliography{aaai23}

\end{document}